\documentclass[a4paper,conference]{IEEEtran}
\ifCLASSINFOpdf
\else
\fi

\usepackage{bm}
\usepackage{graphicx}
\usepackage{multirow}
\usepackage{makecell}
\usepackage{array}
\usepackage{booktabs}
\usepackage{footnote}
\usepackage{threeparttable}

\newcommand{\etal}{\textit{et al}.}

\newcommand{\eg}{\textit{e}.\textit{g}.}
\newcommand{\tabincell}[2]{\begin{tabular}{@{}#1@{}}#2\end{tabular}}

\begin{document}
%
\title{HMFlow: Hybrid Matching Optical Flow Network for Small and Fast-Moving Objects}

\author{\IEEEauthorblockN{Suihanjin Yu, Youmin Zhang, Chen Wang, Xiao Bai, Liang Zhang}
\IEEEauthorblockA{School of Computer Science and Engineering\\
Beihang University\\
Beijing, China\\
Email: \{fakecoderemail, youmi, wangchenbuaa, baixiao, liang.z\}@buaa.edu.cn}
\and
\IEEEauthorblockN{Edwin R. Hancock}
\IEEEauthorblockA{Department of Computer Science\\
University of York\\
York, UK\\
Email: erh@cs.york.ac.uk}}


%


\maketitle

\begin{abstract}
In optical flow estimation task, coarse-to-fine (C2F) warping strategy is widely used to deal with the large displacement problem and provides efficiency and speed. However, limited by the small search range between the first images and warped second images, current coarse-to-fine optical flow networks fail to capture small and fast-moving objects which disappear at coarse resolution levels. To address this problem, we introduce a lightweight but effective Global Matching Component (GMC) to grab global matching features. We propose a new Hybrid Matching Optical Flow Network (HMFlow) by integrating GMC into existing coarse-to-fine networks seamlessly. Besides keeping in high accuracy and small model size, our proposed HMFlow can apply global matching features to guide the network to discover the small and fast-moving objects mismatched by local matching features. We also build a new dataset, named Small and Fast-Moving Chairs (SFChairs), for evaluation. The experimental results show that our proposed network achieves considerable performance, especially at regions with small and fast-moving objects.
\end{abstract}


%
\IEEEpeerreviewmaketitle

\section{Introduction}
Optical flow estimation plays an important role in many computer vision tasks, such as video object segmentation~\cite{ochs2014segmentation, tsai2016video}, action recognition~\cite{wang2013action, simonyan2014two}, and autonomous driving~\cite{menze2015object, costante2018ls}.

Traditional methods typically estimate optical flow by energy minimization in a coarse-to-fine (C2F) framework \cite{glazer1987hierarchical, anandan1989computational, bergen1992hierarchical, brox2004high}. The early deep learning based end-to-end optical flow networks \cite{dosovitskiy2015flownet, ilg2017flownet} are based on encoder-decoder architecture, which possess strong flexibility with large size of model parameters, causing high computing cost. Recently, the coarse-to-fine architecture networks~\cite{ranjan2017optical, hui2018liteflownet, sun2018pwc} readopt traditional coarse-to-fine warping strategy~\cite{glazer1987hierarchical} to provide accuracy and speed. These networks achieve high performances with relatively small model sizes, while they also inherit the problem of capturing the small and fast-moving objects from traditional coarse-to-fine methods.

The main problem lies in the contradiction between limited search range of warping based local matching and large displacement of small objects. To deal with large displacement problem efficiently, the coarse-to-fine warping strategy only sets a very limited search range at all resolution levels and matches between the first images and warped second according to up-sampled flows estimated at previous resolution levels. For a small and fast-moving object whose relative motion is larger than its own scale \cite{brox2011large}, if it disappears at the low-resolution levels, its location in the warped high-resolution image will shift with the background. Once it offsets out of the local search range, the mismatching occurs. And the flow estimation will be wrong at these regions at the subsequent high-resolution levels.

\begin{figure}[t]
\begin{center}
\includegraphics[width=\linewidth]{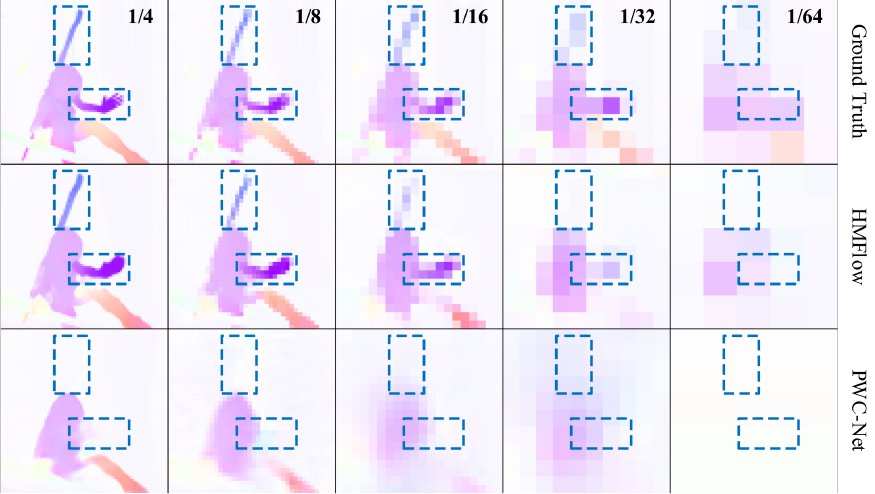}
\end{center}
   \caption{The Flow Spatial Pyramid. The swinging arm and slender stick in blue boxes are small and fast-moving objects.}
\label{fig:example}
\end{figure}

To solve the aforementioned problems, in this paper, we propose an approach to enhance the performance of coarse-to-fine networks at small and fast-moving object regions. Meanwhile, the proposed network only introduces a few network parameters but keep both high accuracy and computing efficiency. The key idea is to guide the coarse-to-fine network by global matching information for long-distance matching. A lightweight Global Matching Component (GMC) is proposed to calculate global matching features without image warping. To achieve our goal, we build Hybrid Matching Optical Flow Network (HMFlow) by integrating GMC into coarse-to-fine network. In conjunction with the limited local matching features of coarse-to-fine network, the global matching features of GMC can reduce the reliance of up-sampled flows, and guide the network discover the flows of small and fast-moving objects at the resolution levels which can distinguish them. And the high-quality warping based local matching features can make the HMFlow keep high accuracy. Fig.~\ref{fig:example} demonstrates the performance of HMFlow compared to PWC-Net~\cite{sun2018pwc}. The flows of swinging arm and slender stick, which are missed by PWC-Net, can be recovered by HMFlow as they appear at $1/16$ resolution level.

We build a new dataset for small and fast-moving objects, named Small and Fast-Moving Chairs (SFChairs). The scenes in SFChairs include small and fast-moving foreground objects and slow-moving background. The profiles and motions of every foreground objects are recorded for quantitative and qualitative evaluations. The effectiveness of network is borne out by experiments on this dataset.

We summarize our contributions as follows. First, we propose a lightweight but effective Global Matching Component (GMC) to produce global matching features that can cover the motion range of small and fast-moving objects. Second, we propose a Hybrid Matching Optical Flow Network (HMFlow), which integrates GMC to capture the small and fast-moving objects. Our proposed network still keeps a lightweight model size and high accuracy. Third, we build a specific dataset, SFChairs, for flow estimation evaluation, especially for small and fast-moving object regions. Experiments based on this dataset show both quantitative and qualitative results.

\section{Related Work}
\textbf{Traditional Methods.} The work of Horn and Schunck~\cite{horn1981determining} is the pioneer for optical flow estimation. This method estimates the optical flow by optimizing an energy function based on brightness constancy and spatial smoothness assumptions. Several works \cite{black1996robust, shulman1989regularization, bruhn2005towards} follow this pipeline.

To efficiently solve the large displacement problem, several works \cite{glazer1987hierarchical, anandan1989computational, bergen1992hierarchical, brox2004high} using coarse-to-fine (C2F) warping strategy are introduced. This strategy estimates flows in a multi-resolution spatial pyramid. An initial coarse flow is estimated and rectified in higher resolution by local matching between the first image and the warped second image according to coarser flow. However this scale-based methods is doom to fail in small and fast-moving objects due to their reliance on low-resolution estimation results. Xu \etal~\cite{xu2012motion} use the SIFT descriptors to match these objects and extend the initial flow accordingly. Thomas \etal~\cite{brox2009large, brox2011large} introduce sparse discriptor matching to guide the variational method to address this problem. Hu~\etal~\cite{hu2016efficient} estimates the flow of small and fast-moving objects by setting and matching new pixel seeds at different levels. All of the above solutions rely on the introduction of global matching information. But they are not differentiable and cannot be easily trained in an end-to-end manner. Our GMC is inspired by them and realized by neural convolutional network.

\textbf{Deep Learning Methods.} FlowNet~\cite{dosovitskiy2015flownet} and FlowNet 2.0~\cite{ilg2017flownet} are encoder-decoder architecture based end-to-end optical flow networks. The range of corresponding features they search for is the size of receptive field. By first encoding and then decoding, they expand receptive field to whole images and achieve global matching without warping according to low-resolution flows. The encoder-decoder architecture performs better than coarser-to-fine at dealing with flow estimation of small and fast-moving objects~\cite{dosovitskiy2015flownet}. However, it also has to train a model with very large size.

Based on coarse-to-fine architecture, recent end-to-end optical flow networks~\cite{ranjan2017optical, hui2018liteflownet, sun2018pwc} can get the accuracy on par with encoder-decoder networks but extremely reduce the model size. SPyNet warps the second image according to coarser flows, and then refined the estimated flows by convoluting the first and the warped image. Rather than warping on image, LiteFlowNet, PWC-Net and their varieties~\cite{hui20liteflownet2, Sun2018:Model:Training:Flow}, within a pretty small search range (e.g., only 4 neighboring pixels~\cite{sun2018pwc, hui2018liteflownet}) at each resolution level, calculate their matching cost with cost volume upon the first and warped image features. However, all these models fail on small and fast-moving objects because of their limited searching range within coarse-to-fine architecture \cite{sun2018pwc, lu2020devon}. Compared with above-mentioned methods, our HMFlow with lightweight GMC can achieve high accuracy while keeping a small size of the model.

Devon~\cite{lu2020devon} avoids using spatial pyramid and warping, and gives much more accurate estimation of the small and fast-moving objects than \cite{hui2018liteflownet, sun2018pwc}. But  it maintains full resolution at all stages and causes huge memory consumption. Our HMFlow with complete coarse-to-fine architecture is more memory friendly and can achieve higher accuracy.

\textbf{Datasets.} The common public datasets for the evaluation of optical flow methods, such as FlyingChairs~\cite{dosovitskiy2015flownet}, FlyingThings3D~\cite{mayer2016large}, MPI Sintel~\cite{butler2012naturalistic}, KITTI 2012~\cite{geiger2012we}, and KITTI 2015~\cite{menze2015object}, just concentrate on general problems of optical flow estimation. Some specialized datasets are built for special problems. Flying Vehicles with Rain (FVR)~\cite{li2017robust} is an optical flow dataset for rainy scenes. RoamingImages~\cite{janai2018unsupervised} is for multi-frame optical flow with occlusions. However, there is no specific dataset for evaluating the performance on regions with small and fast-moving objects. In this paper, we build a new synthetic dataset, by which the quantitative and qualitative evaluation for this problem can be done.

\section{Approach}
Our approach solves the problem of capturing the small and fast-moving objects of end-to-end coarse-to-fine (C2F) networks. We introduce global matching features to mitigate the short of limited search range of C2F network's warping based local matching. We design a lightweight Global Matching Component (GMC) to learn global matching features. We combine the global and local matching features of GMC and C2F network into hybrid matching features, and build Hybrid Matching Optical Flow Network (HMFlow). The hybrid matching features can guide network to capture these missed objects and keep network's high accuracy.

\subsection{Matching And Estimating}\label{sec:match_estimate}
The key of optical flow estimation is matching. From this perspective, the end-to-end optical flow networks can be divided into two parts called the feature matching part and the flow estimating part. The former computes matching features, and the latter estimates optical flow with matching features.

The feature matching part $M$ is used to calculate the matching features. We define this part as follow:
\begin{equation}
\bm{{\rm m}}^0,\cdots,\bm{{\rm m}}^l,\cdots,\bm{{\rm m}}^L,\bm{{\rm u}}^0,\cdots,\bm{{\rm u}}^l,\cdots,\bm{{\rm u}}^L=M(\bm{{\rm i}}_1,\bm{{\rm i}}_2)
\label{equ:feat_match_part}
\end{equation}
where $\bm{{\rm i}}_1$ and $\bm{{\rm i}}_2$ are two consecutive images, $L$ is the max spatial pyramid level, the superscript ${\cdot}^l$ means the feature's resolution is $1/2^l$ of $\bm{{\rm i}}_1$ and $\bm{{\rm i}}_2$, $\bm{{\rm m}}^l$ and $\bm{{\rm u}}^l$ are the output matching feature, which associates pixels in $\bm{{\rm i}}_1$ with their corresponding pixels in $\bm{{\rm i}}_2$, and unmatching feature at $1/2^l$ resolution level.

The flow estimating part $E$ estimates flows according to the outputs of $M$. We define this part at $1/2^l$ resolution level as follows:
\begin{equation}
\bm{{\rm f}}^l=E^l(\bm{{\rm m}}^l,\bm{{\rm u}}^l)
\label{equ:flow_estimate_part}
\end{equation}
where $\bm{{\rm f}}^l$ is estimated flow at $1/2^l$ resolution level.

\begin{figure}[t]
\begin{center}
\includegraphics[width=\linewidth]{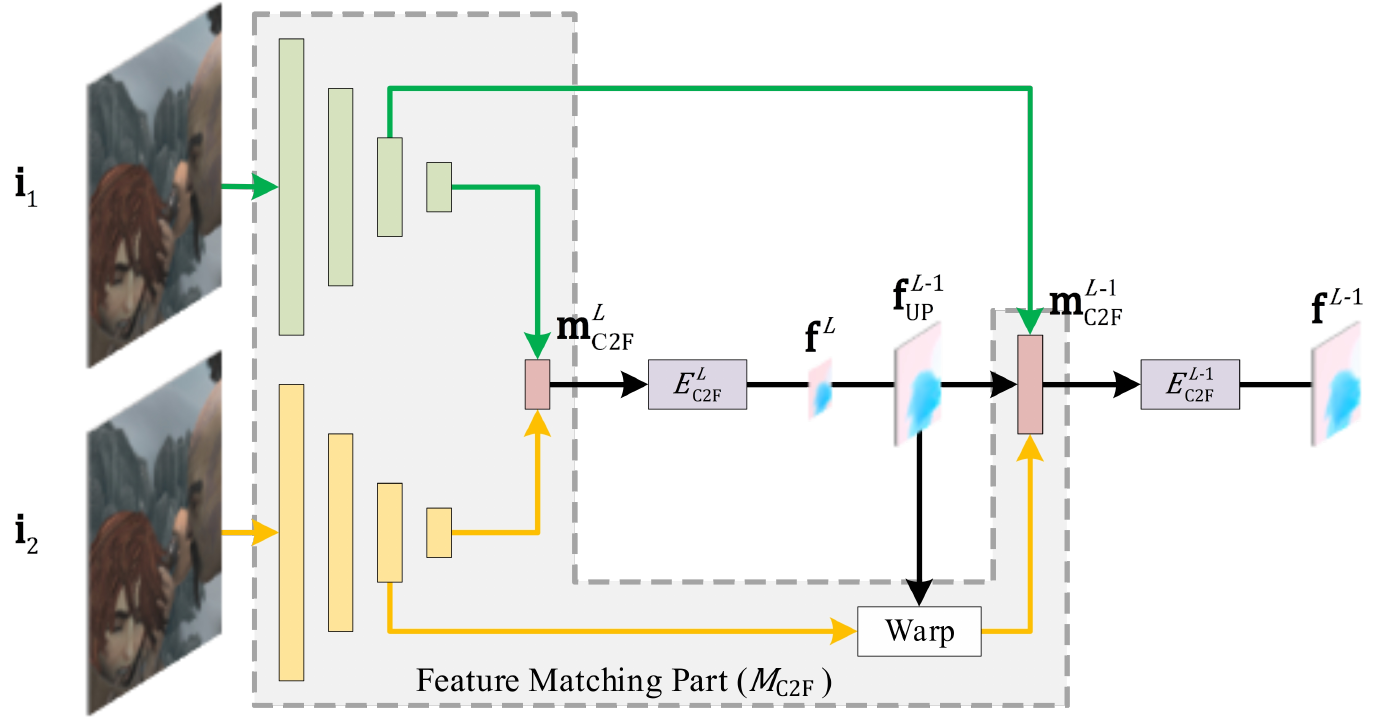}
\end{center}
   \caption{The Function Partition of C2F Network. The feature matching part $M_{\rm C2F}$ is in the grey region, the flow estimating part $E_{\rm C2F}$ is the purple blocks.}
\label{fig:c2f_net}
\end{figure}

The function partition of C2F network is shown in Fig.~\ref{fig:c2f_net}. The feature matching part $M_{\rm C2F}$ in the grey region includes a Siamese network filled by green and yellow, and red matching layers. The matching layers, like warp~\cite{ranjan2017optical,hui2018liteflownet} and cost volume~\cite{hui2018liteflownet,sun2018pwc} produce matching features $\bm{{\rm m}}^l_{\rm C2F}$. The flow estimating part $E^l_{\rm C2F}$ in the purple block is the convolution layers between the matching feature and estimated optical flow at $1/2^l$ resolution level. By convolutional neural network, $E^l_{\rm C2F}$ can directly obtain the refined flow $\bm{{\rm f}}^l$ according to the warping based local matching feature $\bm{{\rm m}}^l_{\rm C2F}$ and the up-sample flow $\bm{{\rm f}}^l_{\rm UP}$ without the need to estimate the residual flow first \cite{sun2018pwc, Sun2018:Model:Training:Flow}.

For C2F networks, the matching features $\bm{{\rm m}}^l_{\rm C2F}$ are the only matching information for the corresponding flow estimating part $E^l_{\rm C2F}$. For an $L$-level pyramid setting, C2F networks only need a partial matching cost with a limited search radius of $r$ pixels. A one-pixel motion at the top level corresponds to $2^{L-1}$ pixels at the full resolution images. Thus the $r$ are set to be small.

\begin{figure}
\begin{center}
\includegraphics[width=\linewidth]{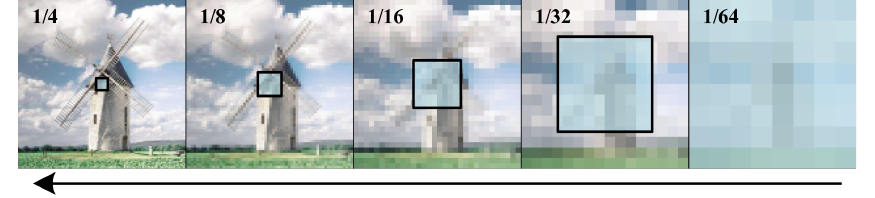}
\end{center}
   \caption{The Local Search Range of C2F Network. In this sample, the resolution of inputs is $512\times512$, the search radius is $r=4$ pixels \cite{sun2018pwc}, the blue region is the search range relative to inputs' resolution at each resolution level. The matching order of network is in arrow direction.}
\label{fig:search_range}
\end{figure}

Fig.~\ref{fig:search_range} demonstrates that the range that $\bm{{\rm m}}^l_{\rm C2F}$ can cover decreases as resolution increases. Hence matching process must depend on the warped features according to up-sampled flows to deal with large displacement. This design comes from traditional methods and leads to the same problem of capturing the small and fast-moving objects.

\subsection{Global Matching Component (GMC)}
To capture the small and fast-moving objects, traditional C2F methods introduce new global sparse descriptor matching to discover the missed objects. But they are not differentiable and cannot be easily trained in an end-to-end manner. The image-guide filter networks~\cite{li2016deep, li2019joint} use image features to guide networks to discover more details. But it’s not enough to recover the motion of the objects moving alone, because these image features can just provide with structure information, rather than the matching information. The encoder-decoder optical flow networks~\cite{dosovitskiy2015flownet, ilg2017flownet} can produce global matching features by expanding receptive field, but their model sizes are too large to be used for small optical flow networks.

\begin{table}
\renewcommand{\arraystretch}{1.3}
\caption{Global Matching Component (GMC)}
\label{table_example}
\centering
\begin{threeparttable}
\begin{tabular}{p{3pt}|l|c|c|c|c|c}
\hline
& \bfseries Layer & \bfseries Kernel & \bfseries Str. & \bfseries Ch. & \bfseries Input & \bfseries Output\\
\hline\hline
\multirow{11}{*}{\rotatebox{90}{\makecell{Encoding}}} & Conv1A & $7\times7$ & $2$ & $c_1$ & $\bm{{\rm i}}_1||\bm{{\rm i}}_2$ & $\bm{{\rm c}}^1_{\rm A}$\\
& Conv1B & $7\times7$ & $1$ & $c_1$ & $\bm{{\rm c}}^1_{\rm A}$ & $\bm{{\rm c}}^1_{\rm B}$\\
& Conv2A & $5\times5$ & $2$ & $c_2$ & $\bm{{\rm c}}^1_{\rm B}$ & $\bm{{\rm c}}^2_{\rm A}$\\
& Conv2B & $5\times5$ & $1$ & $c_2$ & $\bm{{\rm c}}^2_{\rm A}$ & $\bm{{\rm c}}^2_{\rm B}$\\
& Conv3A & $3\times3$ & $2$ & $c_3$ & $\bm{{\rm c}}^2_{\rm B}$ & $\bm{{\rm c}}^3_{\rm A}$\\
& Conv3B & $3\times3$ & $1$ & $c_3$ & $\bm{{\rm c}}^3_{\rm A}$ & $\bm{{\rm c}}^3_{\rm B}$\\
\cline{2-7}
& $\cdots$ & $\cdots$ & $\cdots$ & $\cdots$ & $\cdots$ & $\cdots$\\
\cline{2-7}
& Conv$l$A & $3\times3$ & $2$ & $c_{l}$ & $\bm{{\rm c}}^{l-1}_{\rm B}$ & $\bm{{\rm c}}^l_{\rm A}$\\
& Conv$l$B & $3\times3$ & $1$ & $c_l$ & $\bm{{\rm c}}^l_{\rm A}$ & $\bm{{\rm c}}^l_{\rm B}$\\
\cline{2-7}
& $\cdots$ & $\cdots$ & $\cdots$ & $\cdots$ & $\cdots$ & $\cdots$\\
\cline{2-7}
& Conv$L$A & $3\times3$ & $2$ & $c_{L}$ & $\bm{{\rm c}}^{L-1}_{\rm B}$ & $\bm{{\rm c}}^L$\\
\hline
\multirow{6}{*}{\rotatebox{90}{\makecell{Decoding}}} & Match$L$A & $3\times3$ & $1$ & $c_{L}$ & $\bm{{\rm c}}^{L}$ & $\bm{{\rm m}}^L_{\rm G}$\\
& Deconv$L$-$1$ & $4\times4$ & $2$ & $c_{L-1}$ & $\bm{{\rm m}}^L_{\rm G}$ & $\bm{{\rm d}}^{L-1}$\\
& Match$L$-$1$ & $3\times3$ & $1$ & $c_{L-1}$ & \tabincell{c}{$\bm{{\rm d}}^{L-1}||$\\$\bm{{\rm c}}^{L-1}_{\rm B}||\bm{{\rm f}}^{L-1}_{\rm UP}$} & $\bm{{\rm m}}^{L-1}_{\rm G}$\\
\cline{2-7}
& $\cdots$ & $\cdots$ & $\cdots$ & $\cdots$ & $\cdots$ & $\cdots$\\
\cline{2-7}
& Deconv$l$ & $4\times4$ & $2$ & $c_{l}$ & $\bm{{\rm m}}^{l-1}_{\rm G}$ & $\bm{{\rm d}}^{l}$\\
& Match$l$ & $3\times3$ & $1$ & $c_{l}$ & $\bm{{\rm d}}^{l}||\bm{{\rm c}}^{l}_{\rm B}||\bm{{\rm f}}^{l}_{\rm UP}$ & $\bm{{\rm m}}^l_{\rm G}$\\
\hline
\end{tabular}
\label{tab:gmc}
\begin{tablenotes}
\footnotesize
\item[a] The \textbf{Str.} and \textbf{Ch.} indicate the Stride and Output Channels of convolution layers.
\end{tablenotes}
\end{threeparttable}
\end{table}

Inspired by the above work, we design a Global Matching Component (GMC). The GMC is a lightweight U-Net~\cite{ronneberger2015u} encoder-decoder network, which belongs to feature matching part $M_{\rm GMC}$ functionally. It focuses on providing $E^l_{\rm C2F}$ of C2F network with supplementary global matching features $\bm{{\rm m}}^l_{\rm G}$ by small calculation cost. And $\bm{{\rm m}}^l_{\rm G}$ will guide $E^l_{\rm C2F}$ to discover the missed small and fast-moving objects.

\begin{figure}
\begin{center}
\includegraphics[width=\linewidth]{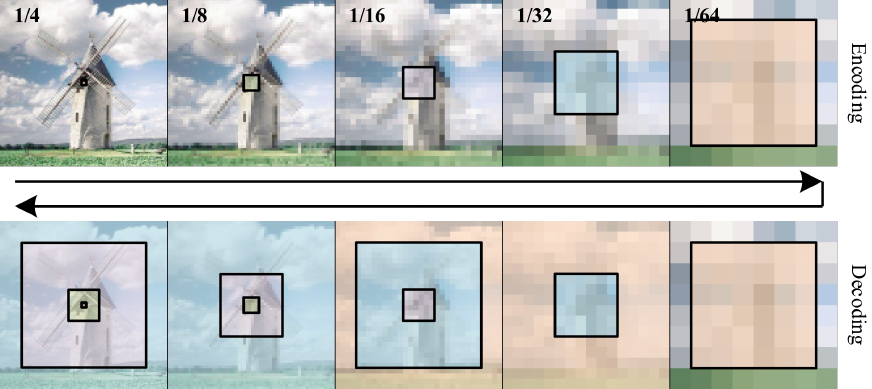}
\end{center}
   \caption{The Theoretical Receptive Field of GMC. In this sample, the resolution of inputs is $512\times512$, the colored region is the search range relative to inputs' resolution at each resolution level. The convolution order is in arrow direction.}
\label{fig:receptive_field}
\end{figure}

The basic architecture of GMC is described in TABLE~\ref{tab:gmc}. The $\bm{{\rm i}}_1||\bm{{\rm i}}_2$ is two consecutive images concatenated in channel dimension. And the $\bm{{\rm m}}^l_{\rm G}$ is the output global matching feature of GMC at $1/2^l$ resolution level. Same as encoder-decoder optical flow networks, the search range of GMC is the receptive fields of its convolution layers. As Fig.~\ref{fig:receptive_field} shown, the theoretical receptive field of GMC expands by encoding, and further expands through skip-connection in decoding. By this way, the search range of $\bm{{\rm m}}^l_{\rm G}$ increases with resolution to whole images, and keep detail information at the same time. This process avoids the dependence of warped features according to up-sampled flows.

\subsection{Hybrid Matching Optical Flow Network (HMFlow)}
\label{sec:hmflow}

\begin{figure}
\begin{center}
\includegraphics[width=\linewidth]{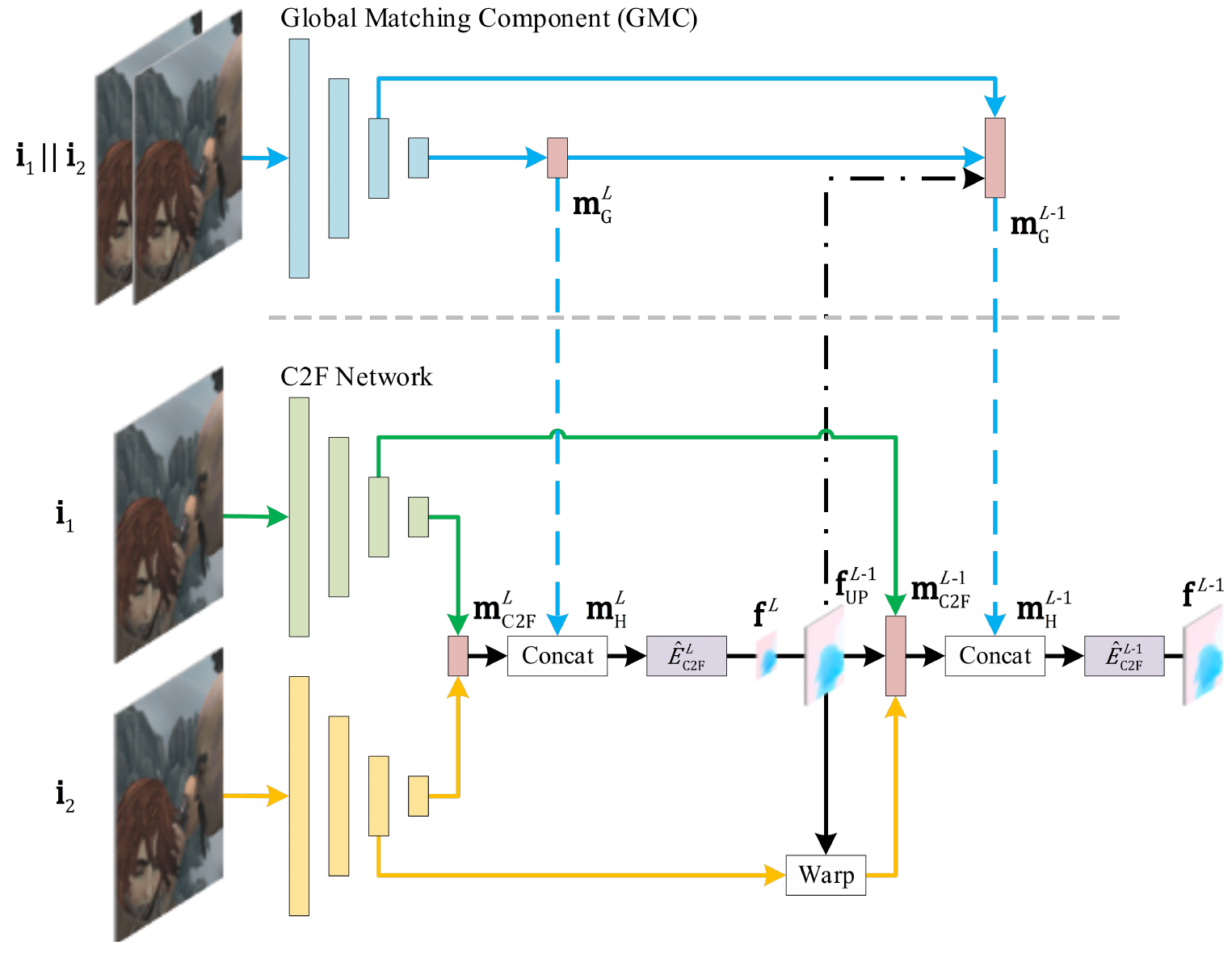}
\end{center}
   \caption{Hybrid Matching Optical Flow Network (HMFlow). The GMC and C2F network are above and below the grey dotted line respectively.}
\label{fig:hmflow}
\end{figure}

Taking advantage of large search range of GMC's global matching features and high-quality of C2F network's local matching features, we build new Hybrid Matching Optical Flow Network (HMFlow) to capture the small and fast-moving objects and keep network's high accuracy and small model size. This new network in Fig.~\ref{fig:hmflow} includes two part, a GMC and a C2F network. HMFlow can be built according to existing C2F network in three steps, (i) network dividing, (ii) component building, and (iii) network integrating.

\textbf{Network Dividing.} Selecting a C2F network and setting its max spatial pyramid level to be $L_{\rm C2F}$. Dividing this network into feature matching part $M_{\rm C2F}$ and flow estimating part $E^l_{\rm C2F}$ according to \ref{sec:match_estimate}.

\textbf{Component Building.} Setting GMC's max spatial pyramid level to be $L_{\rm G}=L_{\rm C2F}$. Setting GMC's number of channels $c_{l}$ according to the corresponding one in the Siamese network of selected C2F network.

\textbf{Component Integrating.} The matching features of GMC and selected C2F network are concatenated in channel dimension to hybrid matching features $\bm{{\rm m}}^l_{\rm H}$:
\begin{equation}
\bm{{\rm m}}^l_{\rm H}=\bm{{\rm m}}^l_{\rm G}||\bm{{\rm m}}^l_{\rm C2F}
\label{equ:hybrid_match}
\end{equation}
And then, the number of channels of $E^l_{\rm C2F}$ are rectified to fit $\bm{{\rm m}}^l_{\rm H}$. The $\bm{{\rm m}}^l_{\rm H}$ are inputed to the rectified $\hat{E}^l_{\rm C2F}$ in place of $\bm{{\rm m}}^l_{\rm C2F}$:
\begin{equation}
\bm{{\rm f}}^l=\hat{E}^l_{\rm C2F}(\bm{{\rm m}}^l_{\rm H},\bm{{\rm u}}^l_{\rm C2F})
\label{equ:rectified_c2f}
\end{equation}
Finally, the up-sampled flows $\bm{{\rm f}}^l_{\rm UP}$ are sent to Match$l$ of GMC.

In HMFlow, $\bm{{\rm m}}^l_{\rm G}$ with global search range can guide $\hat{E}^l_{\rm C2F}$ to discover the small and fast-moving objects dismissed by $\bm{{\rm m}}^l_{\rm C2F}$, and $\bm{{\rm m}}^l_{\rm C2F}$ with high-quality can make the network keep high accuracy. Meanwhile, $\bm{{\rm f}}^l_{\rm UP}$ can assist the GMC in more accurate global matching. By this way, HMFlow can realize bi-directional enhancement of GMC and C2F network.

\begin{figure}
\begin{center}
\includegraphics[width=\linewidth]{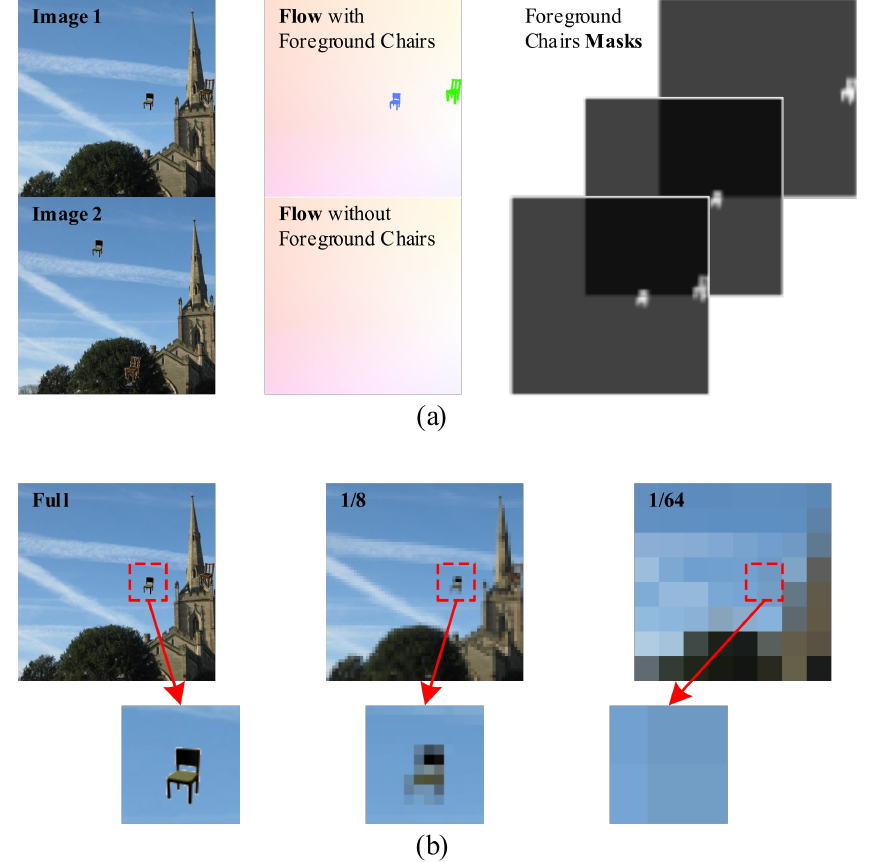}
\end{center}
   \caption{SFChairs Dataset. (a) is a sample with images, optical flow ground truth, and masks for foreground chairs in SFChairs. (b) indicates the foreground chair at different resolution levels, the chair disappears at the lowest resolution level.}
\label{fig:dataset}
\end{figure}

\begin{figure*}[!t]
\begin{center}
\includegraphics[width=\linewidth]{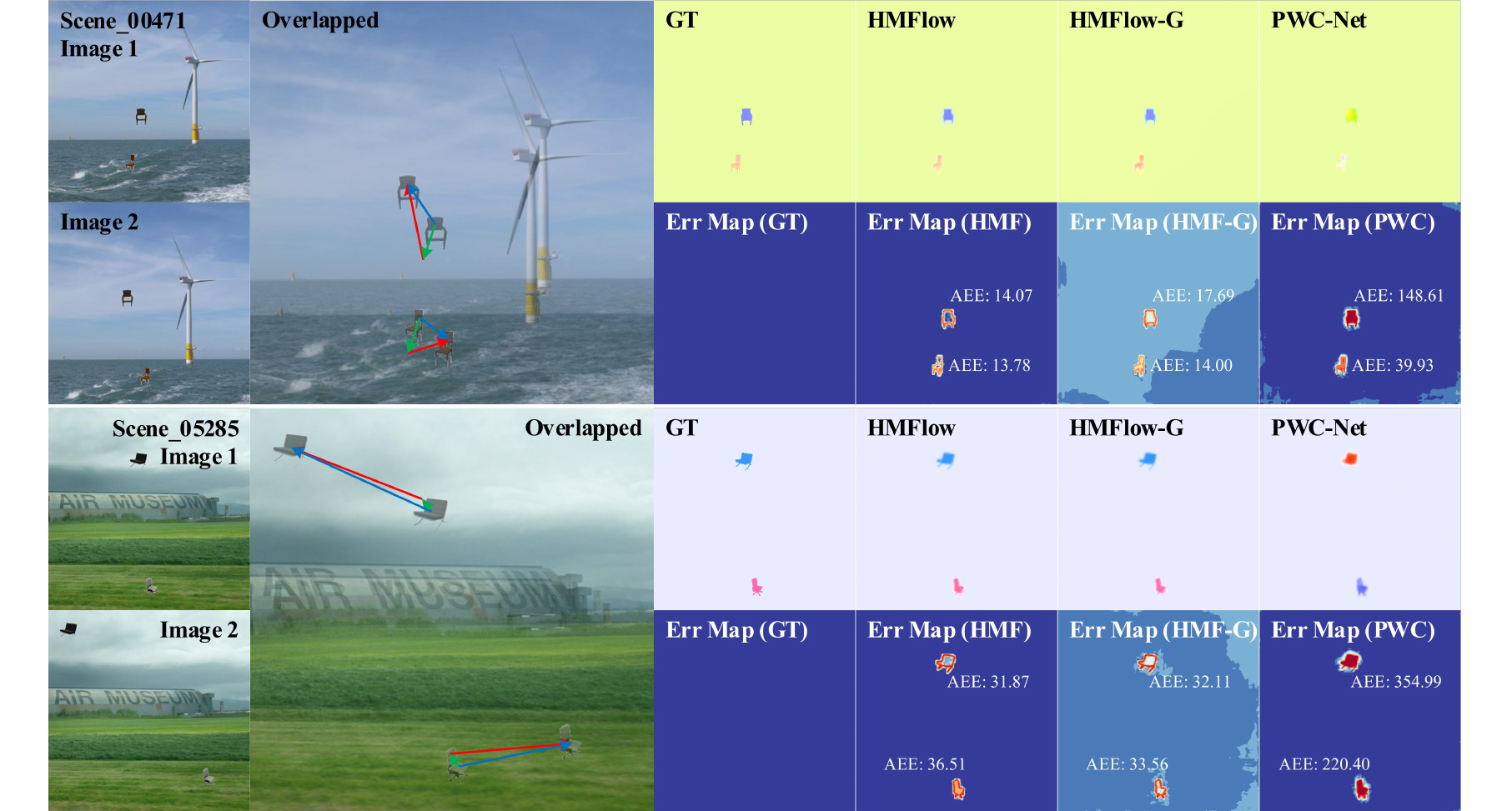}
\end{center}
   \caption{The Results And Error Maps on The Test Set of SFChairs. The absolute and relative trajectory of foreground objects and the motions of background behind the objects are marked by blue, red, and green arrow line in the overlapped images. The AEEs of foreground chairs are marked in the error maps. The HMFlow-G estimates flows with only GMC's global matching features.}
\label{fig:sfchairs}
\end{figure*}

\subsection{Loss}
The HMFlow can be trained using the proposed loss functions. Let $\Theta$ be the set of all learnable parameters in both GMC and C2F network. Let $\bm{{\rm w}}_{\Theta}^l$ denote the estimated flow at the $l$th level, and $\bm{{\rm w}}_{\rm GT}^l$ the corresponding ground truth. At the pre-training stage, we use the same multi-scale training loss as the one used by \cite{dosovitskiy2015flownet}:
\begin{equation}
\mathcal{L}(\Theta)=\sum_{l=l_0}^L\alpha_l\sum_{\bm{{\rm x}}}{|\bm{{\rm w}}_{\Theta}^l-\bm{{\rm w}}_{\rm GT}^l|}_2+\gamma{|\Theta|}_2
\label{equ:loss}
\end{equation}
where ${|\cdot|}_2$ computes the L2 norm of a vector and the second term regularizes parameters of the network. For fine-tuning, we use the robust training loss in \cite{sun2018pwc}:
\begin{equation}
\mathcal{L}(\Theta)=\sum_{l=l_0}^L\alpha_l\sum_{\bm{{\rm x}}}{(|\bm{{\rm w}}_{\Theta}^l-\bm{{\rm w}}_{\rm GT}^l|+\epsilon)}^q+\gamma{|\Theta|}_2
\label{equ:robust_loss}
\end{equation}
where $|\cdot|$ denotes the L1 norm, $q<1$ gives less penalty to outliers, and $\epsilon$ is a small constant.

\section{Small and Fast-Moving Chairs Dataset (SFChairs)}

In order to quantitatively and qualitatively evaluate the performance of optical flow networks in this problem, we build a new dataset for small and fast-moving objects, named Small and Fast-Moving Chairs (SFChairs). Fig.~\ref{fig:dataset} (a) shows a sample in SFChairs. The scenes of this dataset are similar to FlyingChairs, but the scales of the foreground objects and the relative motion between foreground and background are elaborate set according to the definition of small and fast-moving objects in \cite{brox2011large}, which makes all foreground objects become the small and fast-moving objects. Both optical flow ground truth with and without foreground objects is provided. And the pixel-level masks for each foreground object are recorded. We use the chairs from \cite{aubry2014seeing} as foreground and select 2798 images from Places~\cite{zhou2017places} as background. As Fig.~\ref{fig:dataset} (b) shown, the scales of the chairs are less than 64 pixels to guarantee that they are small enough to disappear at the top level of spatial pyramid. SFChairs contains 10,000 examples with a resolution of $512\times512$ that we split into 90\% training set and 10\% test set.

\section{Experiments}
\textbf{Network Details.} We use PWC-Net, a representative end-to-end C2F optical flow networks, as a baseline. HMFlow is built according to the scheme in section~\ref{sec:hmflow}. (i) Network Dividing, PWC-Net's max spatial level is set to be $L_{\rm PWC}=6$, the flow estimating part of PWC-Net $E^l_{\rm PWC}$ is the Optical Flow Estimator at corresponding resolution level, and the feature matching part $M_{\rm PWC}$ is the rest of components. (ii) Component Building, GMC's max spatial pyramid level is set to be $L_{\rm G}=L_{\rm PWC}=6$ and the numbers of channels are set to be $c_1=16$, $c_2=32$, $c_3=64$, $c_4=96$, $c_5=128$, and $c_6=196$ according to PWC-Net's Feature Pyramid Extractor. (iii) Component Integrating, the hybrid matching features $\bm{{\rm m}}^l_{\rm H}$ is made by 3D Cost Volume features of PWC-Net and $\bm{{\rm m}}^l_{\rm G}$ of GMC, the up-sampled flows of PWC-Net are sent to corresponding Match$l$ of GMC.

\textbf{Training Details.} We build and train networks in PyTorch~\cite{paszke2017automatic}. The weights in the training losses, Eq. \ref{equ:loss} and \ref{equ:robust_loss}, are set to be $\alpha_6=0.32$, $\alpha_5=0.08$, $\alpha_4=0.02$, $\alpha_3=0.01$, and $\alpha_2=0.005$. The weight $\gamma$ of L2 penalty is set to be $0.0004$. First, we pre-train HMFlow on FlyingChairs and FlyingThings3D according to learning rate schedule $S_{long}$ and $S_{fine}$ \cite{ilg2017flownet}. Then, we fine-tune network on SFCharis and MPI Sintel.

\subsection{SFChairs}
We fine-tune the pretrained HMFlow and PWC-Net on SFChairs with same learning rate schedule.

\begin{table}
\renewcommand{\arraystretch}{1.3}
\caption{AEEs on SFChairs}
\label{table_example}
\centering
\begin{threeparttable}
\begin{tabular}{l|c|c|c|c|c|c}
\hline
\multirow{2}{*}{\bfseries Models} & \multicolumn{3}{c|}{\bfseries Training Set} & \multicolumn{3}{c}{\bfseries Test Set}\\
\cline{2-7}
& \bfseries All & \bfseries Bg. & \bfseries Obj. & \bfseries All & \bfseries Bg. & \bfseries Obj.\\
\hline\hline
PWC-Net & (0.62) & (0.27) & (64.54) & 0.79 & 0.27 & 87.01\\
HMFlow-G & (0.59) & (0.36) & (45.58) & 0.71 & 0.42 & 56.03\\
HMFlow & (\textbf{0.39}) & (\textbf{0.20}) & (\textbf{36.64}) & \textbf{0.45} & \textbf{0.21} & \textbf{44.34}\\
\hline
\end{tabular}
\label{tab:aee_sfchairs}
\begin{tablenotes}
\footnotesize
\item[a] The \textbf{All}, \textbf{Bg.} and \textbf{Obj.} indicate the AEEs of All image, Background and Foreground Object Regions.
\item[b] The \textbf{HMFlow-G} estimates flows with only GMC's global matching features.
\end{tablenotes}
\end{threeparttable}
\end{table}

\textbf{Quantitative Analysis}. Quantitative evaluation of networks performance based on Average End-Point Error (AEE), which represents the Euclidean distance between the estimated results and the ground truth. In TABLE~\ref{tab:aee_sfchairs}, we compare the AEEs of HMFlow with PWC-Net on SFChairs. The performance of HMFlow has distinct advantage over PWC-Net on both training set and test set. The AEEs of slow-moving background are similar for both networks. However, drawing the global matching features from GMC, HMFlow's AEEs of small and fast-moving foreground objects are about 50\% of PWC-Net. It proves that the improvement effect of HMFlow on this problem in original PWC-Net is obvious.

\textbf{Qualitative Analysis}. Fig.~\ref{fig:sfchairs} compares the results of HMFlow with PWC-Net on the test set of SFChairs. HMFlow can capture the small and fast-moving chairs in the scenes, while their motion is mis-estimated by PWC-Net.

\begin{figure}
\begin{center}
\includegraphics[width=\linewidth]{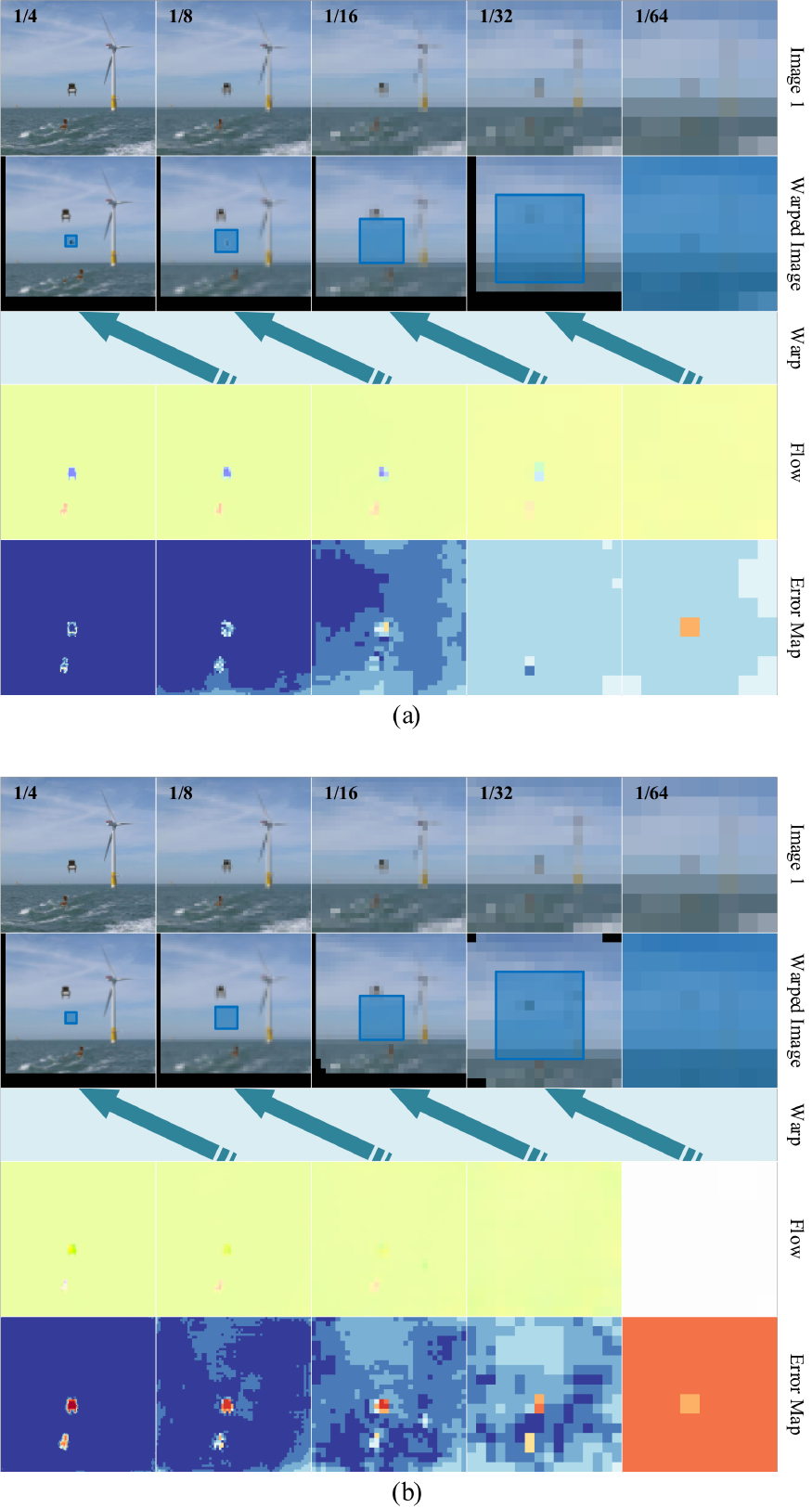}
\end{center}
   \caption{The Results in Spatial Pyramid. (a) and (b) show the results of HMFlow and PWC-Net. The blue regions in the warped images indicate the local search ranges of $\bm{{\rm m}}^l_{\rm C2F}$.}
\label{fig:sfchairs_sp}
\end{figure}

\begin{figure}
\begin{center}
\includegraphics[width=\linewidth]{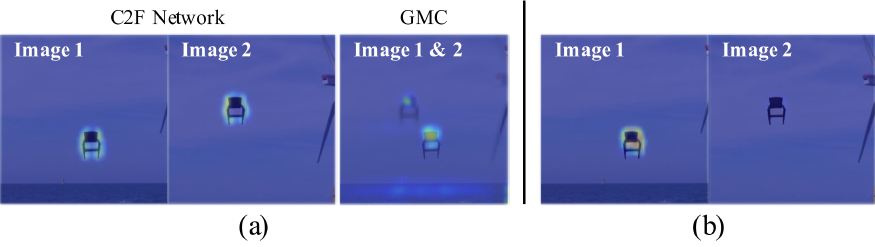}
\end{center}
   \caption{The Saliency Maps. The saliency maps visualise the most concentrated regions of the foreground chair in inputs. (a) and (b) are the maps of HMFlow and PWC-Net.}
\label{fig:sfchairs_sm}
\end{figure}

\begin{figure*}[!t]
\begin{center}
\includegraphics[width=\linewidth]{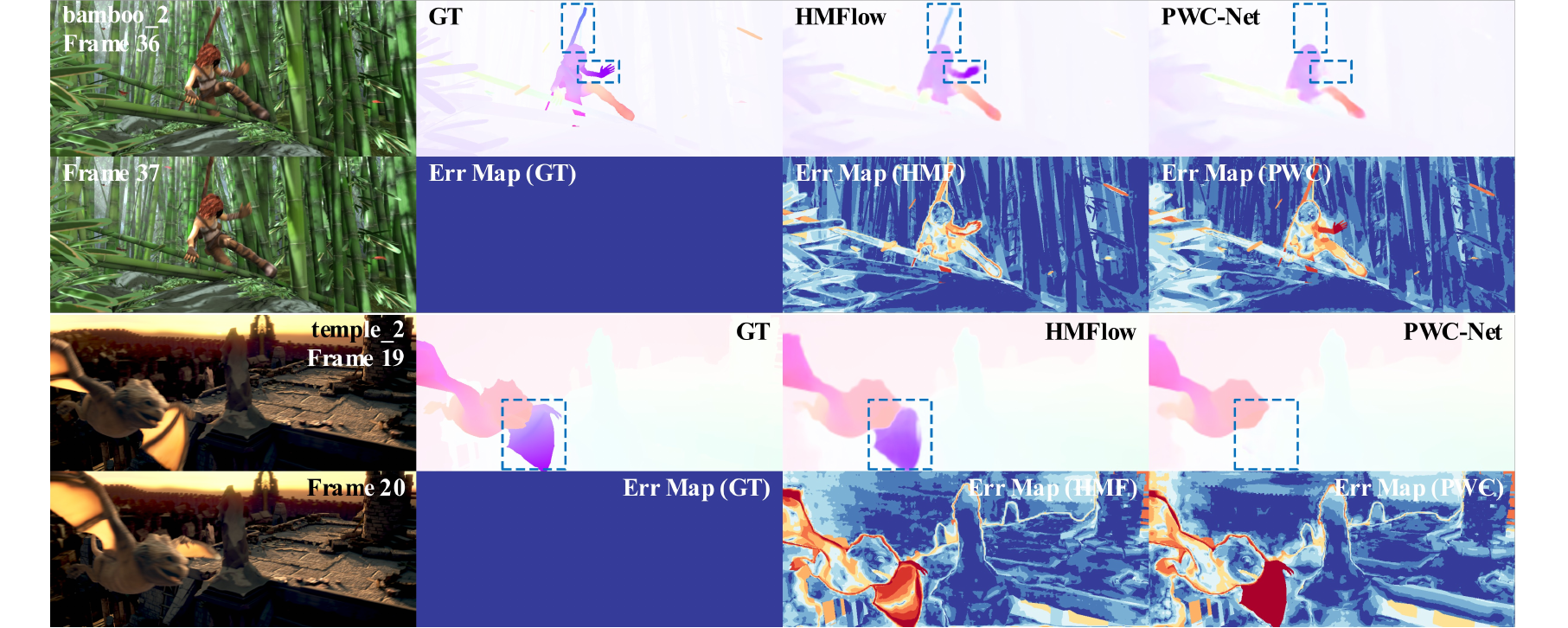}
\end{center}
   \caption{The Results and Error Maps on The Training Set of MPI Sintel. The small and fast-moving objects are marked with blue boxes in optical flows.}
\label{fig:sintel}
\end{figure*}

Fig.~\ref{fig:sfchairs_sp} exhibits how the global matching features of GMC play a role in the spatial pyramid of the C2F network. The chair in the center of the input images can't be seen clearly in the images with $1/64$ and $1/32$ resolution levels. And at $1/16$ resolution level, when the second image is warped according to the up-sampled optical flow, this chair moves with the background and goes beyond the local search range of $\bm{{\rm m}}^4_{\rm C2F}$. PWC-Net's flow estimating part $E^4_{\rm PWC}$ just depends on this local matching feature $\bm{{\rm m}}^4_{\rm PWC}$ whose search range is unable to cover this chair. Hence, this chair's motion will not be able to correctly estimated at current and following resolution levels. In contrast, all $\hat{E}^l_{\rm C2F}$ of HMFlow can get the supplementary global matching information $\bm{{\rm m}}^l_{\rm G}$ from GMC. By $\bm{{\rm m}}^4_{\rm G}$, this chair's flow can be estimated by $\hat{E}^4_{\rm C2F}$ at $1/16$ resolution level, and is warped back to the local search range of subsequent of $\bm{{\rm m}}^3_{\rm C2F}$ and of $\bm{{\rm m}}^2_{\rm C2F}$.

We visualise saliency maps~\cite{simonyan2013deep} to show the divertion of the C2F network's attention before and after the GMC introduction. Fig.~\ref{fig:sfchairs_sm} demonstrates the attentive location of the small and fast-moving chair in the inputs. Limited by local search range, PWC-Net can just pay attention to the chair in the first image in (b). It means that PWC-Net fails to locate this chair from the second image. In contrast, the GMC of HMFlow is able to notice the chair before and after moving at the same time, and guide the C2F network pay attention to the chair in the second image, and match it with the one in the first image.

\textbf{Ablation Study}. To analyze GMC's global matching features, we run an ablation on HMFlow. HMFlow-G removes $M_{\rm C2F}$'s all local matching features $\bm{{\rm m}}^l_{\rm C2F}$, but keeps the unmatching features $\bm{{\rm u}}^l_{\rm C2F}$. All $\hat{E}^l_{\rm C2F}$ of HMFlow-G only use GMC's global matching features $\bm{{\rm m}}^l_{\rm G}$ to estimate optical flows.

In TABLE~\ref{tab:aee_sfchairs}, HMFlow-G's AEEs of foreground objects on training set and test set are about 65\% and 70\% of PWC-Net, while its AEEs of background are higher. The results in Fig.~\ref{fig:sfchairs} also reflect same the same phenomenon. This ablation shows that GMC's global matching features can effectively match the small and fast-moving objects and help network to estimate their motion. But these lightweight global matching features perform obviously poorly in the large smooth regions, \eg background. And our full HMFlow can complement the local and global matching features, and achieves high performance in all regions.

\subsection{MPI Sintel}

We fine-tune the pre-trained HMFlow on MPI Sintel dataset. The robust loss function in Eq.~\ref{equ:robust_loss} is used with $\epsilon=0.01$ and $q=0.4$ \cite{sun2018pwc}. We use the clean and final passes of this training data throughout the fine-tuning process.

\begin{table}
\renewcommand{\arraystretch}{1.3}
\caption{AEEs on MPI Sintel}
\label{table_example}
\centering
\begin{threeparttable}
\begin{tabular}{l|c|c|c|c|c}
\hline
\multirow{2}{*}{\bfseries Methods} & \multicolumn{2}{c|}{\bfseries Training Set} & \multicolumn{2}{c|}{\bfseries Test Set} & \bfseries Size\\ 
\cline{2-5}
& \bfseries Clean & \bfseries Final & \bfseries Clean & \bfseries Final & (million)\\
\hline\hline
FlowNetS~\cite{dosovitskiy2015flownet} & (3.66) & (4.44) & 6.96 & 7.76 & 38.68\\
FlowNetC~\cite{dosovitskiy2015flownet} & (3.78) & (5.28) & 6.85 & 8.51 & 39.18\\
FlowNet2~\cite{ilg2017flownet} & (1.45) & (2.01) & 4.16 & 5.74 & 162.52\\
\hline
SPyNet~\cite{ranjan2017optical} & (3.17) & (4.32) & 6.64 & 8.36 & 1.20\\
LiteFlowNet~\cite{hui2018liteflownet} & (\textbf{1.35}) & (\textbf{1.78}) & 4.54 & 5.38 & 5.37\\
PWC-Net~\cite{sun2018pwc} & (1.70) & (2.21) & 3.86 & 5.13 & 9.37\\
\hline
HMFlow & (1.44) & (2.23) & \textbf{3.21} & \textbf{5.04} & 14.27\\
\hline
\end{tabular}
\label{tab:aee_sintel}
\begin{tablenotes}
\footnotesize
\item[a] The \textbf{Size} indicates networks' number of parameters in million.
\end{tablenotes}
\end{threeparttable}
\end{table}

\textbf{Accuracy and Model Size.} TABLE~\ref{tab:aee_sintel} compares HMFlow with the representative encoder-decoder~\cite{dosovitskiy2015flownet,ilg2017flownet} and C2F~\cite{hui2018liteflownet,sun2018pwc} optical flow networks. The AEEs of HMFlow are lower than all networks participating in the comparison on MPI Sintel test set. Due to the use of GMC, HMFlow's model size is slightly larger than its baseline, PWC-Net. Even though, its parameters are just $1/3$ of FlowNet~\cite{dosovitskiy2015flownet} and $1/11$ of FlowNet 2.0~\cite{ilg2017flownet}. The results show that our HMFlow can take the advantage of C2F network's small model size and relative high accuracy.

\textbf{Small and Fast-Moving Objects.} We qualitatively compare the results of HMFlow with PWC-Net on MPI Sintel training set, which provides ground truth. There are some small and fast-moving objects in the scenes in Fig.~\ref{fig:sintel}, such as the swinging arm, slender stick, and fanning wing. HMFlow can estimate flows of these objects accurately, while PWC-Net dismisses their motion. This shows that our HMFlow is effective for this problem in general situations, instead of overfitting on SFChairs.

\section{Conclusion}
In this paper, we have studied the problem of capturing the small and fast-moving objects in C2F optical flow networks. To address the mismatching of C2F networks' warping based local matching features, we have designed a lightweight Global Matching Component (GMC) for global matching features. We have built new Hybrid Matching Optical Flow Network (HMFlow) by integrating GMC into existing C2F networks. We have built a specialized dataset for small and fast-moving objects, SFChairs. The experiments have proved the HMFlow's effectiveness that our network can solve the problem of capturing small and fast-moving objects with small model size and high accuracy.






\bibliographystyle{IEEEtran}
\bibliography{IEEEabrv,./bib/paper}
%
%
%

\end{document}